\documentclass[10pt,twocolumn,letterpaper]{article}
\usepackage[T1]{fontenc}
\usepackage[pagenumbers]{cvpr}
\usepackage[hidelinks]{hyperref}

\title{PICA: Make $\pi_{0.5}$ Learn to Call Astra for\\
  Reliable and Efficient Robot Manipulation}

\author{%
  Jianli Sun\textsuperscript{1,*}\quad
  Chengxiang Li\textsuperscript{2,*}\quad
  Benquan Tian\textsuperscript{1}\quad
  Xiangdong Li\textsuperscript{3}\\
  Deyi Ji\textsuperscript{4}\quad
  Guo Dabo\textsuperscript{2}\quad
  Bin Tian\textsuperscript{1}\quad
  Yonglin Tian\textsuperscript{1}\\[0.6em]
  {\small\textsuperscript{1}\,Institute of Automation, Chinese Academy of Sciences}\\
  {\small\textsuperscript{2}\,Sanya University}\\
  {\small\textsuperscript{3}\,Institute of Computing Technology, Chinese Academy of Sciences}\\
  {\small\textsuperscript{4}\,University of Science and Technology of China}\\[0.4em]
  {\small\textsuperscript{*}\,Jianli Sun and Chengxiang Li contributed equally and share first authorship.}
}
\date{}

\begin{document}
\maketitle

\begin{abstract}
Vision-language-action (VLA) policies enable fast robot manipulation
but often fail in challenging states, whereas frontier multimodal
reasoning models offer strong recovery at high inference cost. We
present \textbf{PICA}, a selective reasoning--action collaboration
framework that pairs a frozen $\pi_{0.5}$ policy with GPT-6 Astra
through selective assistance. PICA introduces a lightweight, policy-aware CALL
policy that uses internal VLA features and causal execution history to
detect failure risk. To reduce intervention overhead,
experience-guided recovery (EGR) combines configuration-based retrieval
with visual reference selection to guide local corrections and early
VLA resumption. Two independent training regimes support this design: offline learning
uses recorded VLA rollouts, whereas Online Self-Evolution learns CALL
decisions and recovery memory from scratch through outcome-based
first-alarm optimization and intervention feedback. Across four LIBERO suites,
offline-trained PICA improves success from 95.5\% to 99.0\% at 11.46 Hz with 10.99k
tokens per episode, while online self-evolution averages 94.0\% success
during learning on LIBERO-10 from scratch. On a physical SO-101 robot, PICA boosts
success from 70.0\% to 93.3\% with only 13.6\% of actions controlled by
Astra.
\end{abstract}

\section{Introduction}
\label{sec:introduction}

General-purpose multimodal reasoning models such as GPT-6 Astra have
shown strong capabilities in visual reasoning, action generation, and
closed-loop error recovery~\cite{galbotteam2026systematicallyexploringcapabilitiesgpt6,
he2026robotmanipulationgpt6astrabody,yang2026astranomy}. On
RoboLab-120, Astranomy~\cite{yang2026astranomy} achieves a 74.8\%
success rate with an Astra-based agentic policy, outperforming
specialized robot policies. However, continuous control with frontier
reasoning models incurs substantial latency and cost: Astranomy reports
18.56 seconds per direct policy call and \$36.40 per successful episode
in its agentic mode. In contrast, specialized vision-language-action
(VLA) models such as
OpenVLA~\cite{kim2024openvlaopensourcevisionlanguageactionmodel} and
$\pi_{0.5}$~\cite{intelligence2025pi05visionlanguageactionmodelopenworld}
provide fast visuomotor control, but often fail in unfamiliar states or
under accumulated errors. This contrast creates a natural
complementarity: \emph{frontier reasoning models provide strong
deliberation and recovery, while VLAs offer fast routine execution}.

\begin{figure*}[t]
    \centering
    \includegraphics[width=\textwidth]{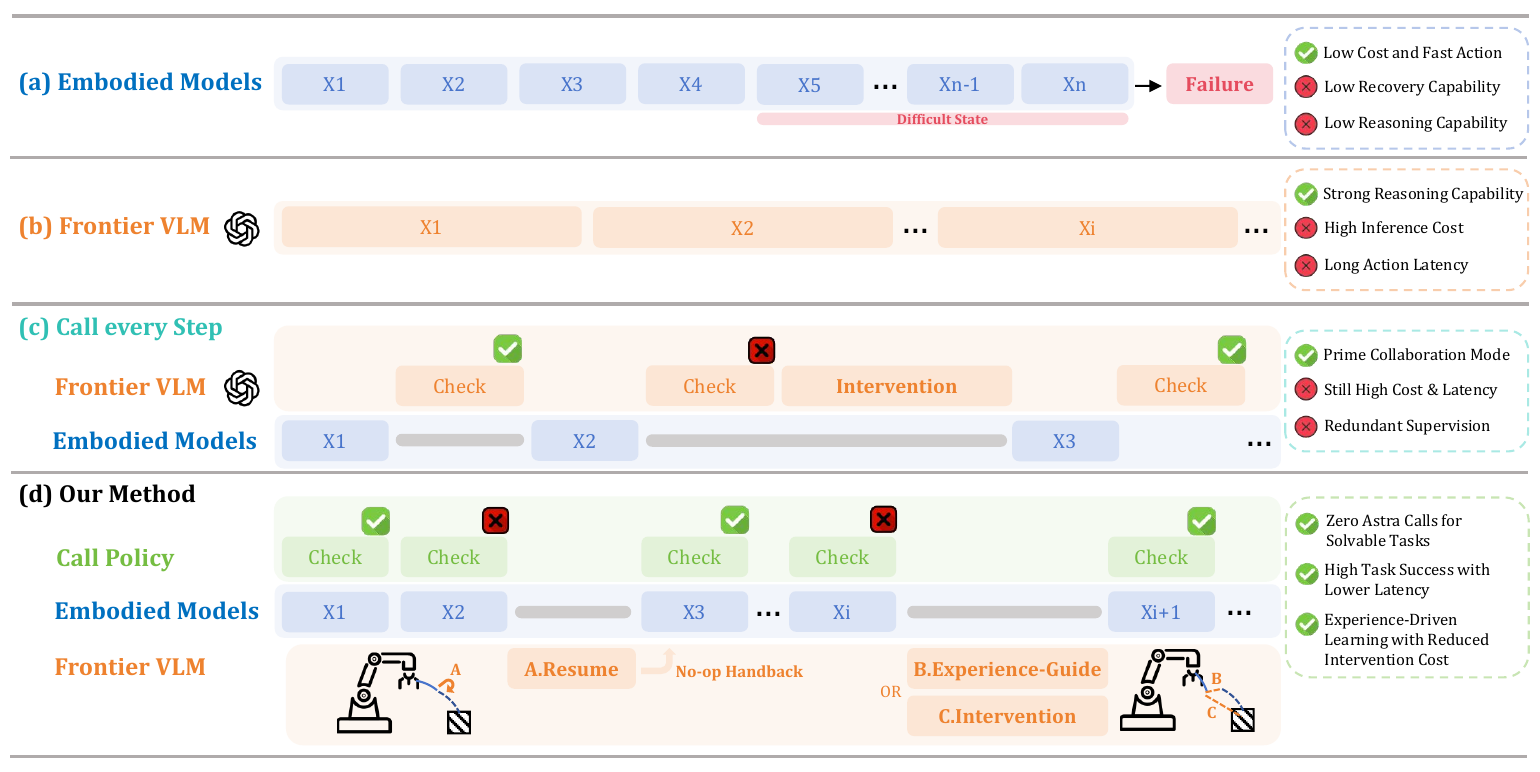}
    \caption{\textbf{Reasoning--action collaboration paradigms.}
    (a) \textbf{Embodied Models}: fast execution, but prone to failure in difficult states.
    (b) \textbf{Frontier VLM}: strong reasoning and recovery, but high cost and latency.
    (c) \textbf{Call Every Step}: continuous VLM supervision incurs redundant overhead.
    (d) \textbf{Our Method (PICA)}: a local CALL policy invokes Astra selectively, which (A) resumes the VLA via no-op handback, (B) applies experience-guided recovery, or (C) performs visual intervention. Online learning further refines CALL decisions and recovery memory.}
    \label{fig:reasoning_action_paradigms}
\end{figure*}

Combining both models is promising~\cite{galbotteam2026systematicallyexploringcapabilitiesgpt6,
yang2026astranomy}, yet efficient collaboration requires more than
simply reducing the fraction of VLM-executed steps. First, reviewing
every VLA proposal incurs heavy reasoning overhead even on solvable
states. Second, once invoked, open-ended VLM control can prolong
intervention when the VLA only needs a brief local correction to
resume. Third, adapting invocation decisions and recovery priors requires
learning from deployment experience. Addressing these issues requires
answering three questions: \emph{when to invoke expensive reasoning,
when to return control to the fast policy, and how to improve both
decisions online} (Fig.~\ref{fig:reasoning_action_paradigms}).

We present \textbf{PICA}, a selective reasoning--action collaboration
framework that pairs a frozen $\pi_{0.5}$ policy with GPT-6 Astra via a
lightweight, policy-aware CALL policy. The CALL policy reuses the VLA's
internal scene-context features, action-intent features, and causal
transition history to detect failure risk without extra VLA forward
passes, keeping the VLA as the default controller.

To shorten external interventions, we introduce \emph{experience-guided
recovery (EGR)} for early policy resumption. PICA maintains a
task-conditioned recovery memory of reference states sampled from
successful VLA rollouts. When an alarm is triggered, configuration-based
retrieval provides up to five same-task references that pass IK
feasibility checks. Astra compares their images with live observations
to select a task-compatible target for bounded IK recovery. It can
also resume a suitable VLA proposal directly or apply visual
intervention when the references are unsuitable. These options support
local correction and early return to fast VLA execution.

PICA supports two independent training regimes. Offline learning trains
the CALL monitor and constructs recovery memory from recorded VLA
rollouts. Alternatively, \emph{Online Self-Evolution} learns the
collaboration layer during deployment~\cite{wang2026adarsiondemandrecursiveselfimprovement,
luo2026recursiveselfimprovementphysicalai}. We optimize a recurrent
request policy over VLA-controlled segments using an outcome-based
first-alarm objective and intervention feedback. The first-alarm
objective marginalizes over first-confirmed-request
voting histories on recorded VLA segments. We also append successful
$\pi_{0.5}$ states to the online recovery memory. Evaluations across
four LIBERO
suites~\cite{liu2023liberobenchmarkingknowledgetransfer} and a physical
SO-101 robot show that PICA achieves high manipulation reliability with
lower reasoning overhead than continuous supervision.

Our main contributions are threefold:
\begin{itemize}
    \item We propose \textbf{PICA}, a selective reasoning--action
    collaboration framework that pairs a frozen VLA with an external
    reasoning model via a lightweight, policy-aware CALL policy.

    \item We develop \textbf{experience-guided recovery (EGR)}, which
    combines configuration-based retrieval, visual reference selection,
    and adaptive recovery modes to support early VLA resumption.

    \item We introduce an \textbf{online self-evolution mechanism}
    combining first-alarm marginalization, intervention feedback,
    and incremental memory growth to learn collaboration from scratch.
\end{itemize}

\section{Related Work}
\label{sec:related_work}

\noindent\textbf{General-Purpose Reasoning for Robot Manipulation.}
Vision-language-action (VLA) models, such as
RT-2~\cite{brohan2023rt2visionlanguageactionmodelstransfer},
OpenVLA~\cite{kim2024openvlaopensourcevisionlanguageactionmodel}, and
$\pi_{0.5}$~\cite{intelligence2025pi05visionlanguageactionmodelopenworld},
have shown strong performance in generalist robot manipulation. At the
same time, general-purpose multimodal reasoning models are increasingly
used for embodied control. Astranomy~\cite{yang2026astranomy} uses
GPT-6 Astra directly as a robot policy, demonstrating strong
manipulation capabilities alongside high inference costs. Galbot
Team~\cite{galbotteam2026systematicallyexploringcapabilitiesgpt6}
further explores Astra-based control and hybrid execution with
pretrained policies. Building on these complementary strengths, we
design a selective collaboration framework that invokes expensive
reasoning selectively and returns control to a fast VLA after local
correction.

\noindent\textbf{Failure Monitoring and Selective Intervention.}
Runtime failure monitoring improves robot reliability by detecting
execution risks early. Sentinel~\cite{pmlr-v270-agia25a} tracks action
consistency and task progress, while
SAFE~\cite{gu2025safemultitaskfailuredetection} predicts failures from
internal VLA features.
VLA-FAIL~\cite{seligmann2026vlafailefficienttaskfailure} identifies
anomalies via feature deviations and action consistency, and
uncertainty-aware policy
steering~\cite{yuan2026actasklearnuncertaintyaware} decides when to act
autonomously or request help. Our online CALL policy combines
outcome-based first-alarm optimization with intervention feedback to
learn selective requests for external reasoning.

\noindent\textbf{Experience-Guided Recovery and Policy Resumption.}
Prior recovery methods handle execution errors using learned corrective
policies or stored experiences. FLARE~\cite{Zhao_2026_CVPR} and
Recova~\cite{liu2026recovaagentguidedfailurerecovery} learn dedicated
recovery behaviors, whereas
Rewind-IL~\cite{zheng2026rewindilonlinefailuredetection} and
RESETTLE~\cite{chen2026resettleroboticrecoverydisagreementtriggered}
use intermediate states or retrieved demonstrations to recover and
resume the base policy. Recent Astra-based
studies~\cite{he2026robotmanipulationgpt6astrabody} and
LEXP~\cite{zeng2026learningphysicalexperiencecontextspaceadaptation}
also show the value of reusing physical experiences for embodied
reasoning. Our EGR module retrieves successful reference states by
robot configuration and uses Astra's visual assessment to select
between reference-guided IK and direct visual correction.

\noindent\textbf{Online Self-Evolution for Embodied Agents.}
Online self-evolution allows embodied agents to improve from deployment
experience~\cite{luo2026recursiveselfimprovementphysicalai}.
RoboCat~\cite{bousmalis2023robocatselfimprovinggeneralistagent} and
RISE~\cite{yang2026riseselfimprovingrobotpolicy} refine robot policies
through self-generated rollouts and policy optimization.
Recova~\cite{liu2026recovaagentguidedfailurerecovery} incorporates
recovery trajectories into subsequent training, while
AdaRSI~\cite{wang2026adarsiondemandrecursiveselfimprovement} combines
skill refinement with selective VLA fine-tuning. Our framework keeps
both foundation models frozen and learns the collaboration layer
through outcome-based CALL policy updates, intervention feedback,
and incremental growth of the recovery memory.

\section{Method}
\label{sec:method}

\begin{figure*}[t]
    \centering
    \includegraphics[width=\textwidth]{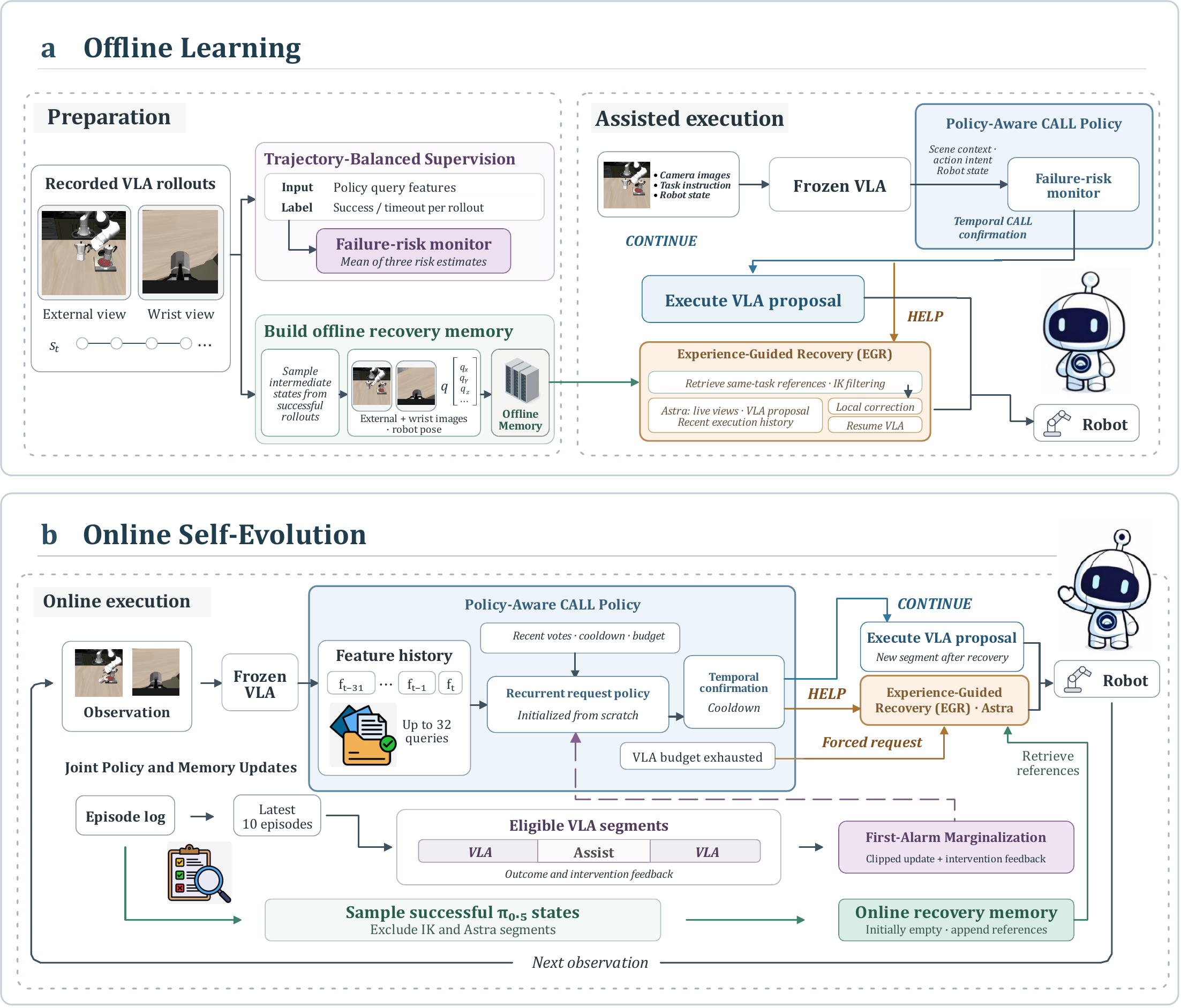}
    \caption{\textbf{PICA: a shared collaboration loop with two alternative training regimes.}
    In both panels, the VLA proposes actions, the CALL policy selects
    \texttt{CONTINUE} or \texttt{HELP}, and Astra uses EGR for local
    correction and early VLA resumption.
    \textbf{(a) Offline learning:} recorded rollouts supervise the
    failure-risk monitor and supply successful reference states before
    assisted execution.
    \textbf{(b) Online Self-Evolution:} a recurrent request policy learns
    from eligible VLA segments and intervention feedback, while
    successful VLA states incrementally populate the recovery memory.}
    \label{fig:method}
\end{figure*}

PICA separates fast action execution from selective reasoning through
two shared components: a policy-aware CALL policy decides \emph{when
to request assistance}, and experience-guided recovery (EGR) decides
\emph{how to select local corrections for VLA resumption}.
We first describe this common execution loop, then introduce two
alternative ways to learn its collaboration layer: offline learning
from recorded rollouts and Online Self-Evolution from deployment.
The two regimes have independent CALL initialization and recovery
memories, while keeping both the VLA and Astra frozen.
In Fig.~\ref{fig:method}a, preparation supplies a monitor and memory
for the assisted execution loop on the right. In
Fig.~\ref{fig:method}b, the upper execution loop supplies
experience to the lower update loop. The regimes differ in training data, CALL architecture, and update
schedule, while sharing selective assistance and policy resumption.

\subsection{Policy-Aware CALL Policy}
\label{sec:call_monitor}

At query step $t$, the frozen VLA predicts an action chunk
$\hat{\mathbf{a}}_t=\pi_{\mathrm{VLA}}(o_t,\ell)$ from observation
$o_t$ and instruction $\ell$. The CALL policy reuses features from
this forward pass, requiring no additional VLA inference. We pool
vision--language prefix tokens into scene-context features
$\mathbf{h}_t^{\mathrm{sem}}$ and flow-matching action tokens into
action-intent features $\mathbf{h}_t^{\mathrm{act}}$. Together with
proprioception $\mathbf{s}_t$, action-chunk statistics $\mathbf{u}_t$,
and causal transition features $\boldsymbol{\delta}_t$, they form
\begin{equation}
\mathbf{f}_t=\big[
\mathbf{h}_t^{\mathrm{sem}};
\mathbf{h}_t^{\mathrm{act}};
\mathbf{s}_t;\mathbf{u}_t;\boldsymbol{\delta}_t
\big].
\label{eq:call_representation}
\end{equation}
The transition features summarize changes in robot motion and internal
representations over the preceding executed action prefix. This
representation links the current scene and intended action to recent
execution progress, allowing CALL decisions to reflect the frozen
policy's own behavior.

\paragraph{Temporal CALL Confirmation.}
The policy aggregates recent assistance decisions to suppress
transient alarms, subject to a post-intervention cooldown. With
\texttt{CONTINUE}, the robot executes the VLA proposal; a confirmed
\texttt{HELP} request invokes Astra. A timeout safeguard separately
forces assistance if the VLA exhausts its execution budget.
Both training regimes use this shared mechanism
(Fig.~\ref{fig:method}); the confirmation threshold and other
implementation settings are specified in Secs.~\ref{sec:exp_setup} and \ref{app:call_settings}.

\subsection{Experience-Guided Recovery}
\label{sec:recovery}

EGR uses successful execution references to guide local corrections
that support VLA resumption. As illustrated by assisted
execution in Fig.~\ref{fig:method}a and external assistance in
Fig.~\ref{fig:method}b, the same recovery procedure serves both regimes.

\paragraph{Reference Memory and Retrieval.}
The memory $\mathcal{M}$ stores intermediate non-terminal states from
successful VLA rollouts. Each entry contains a task ID, external and
wrist RGB images, and robot configuration, including joint angles,
end-effector pose, and gripper width. Offline learning constructs this
memory before deployment; Online Self-Evolution accumulates entries
during execution.

Upon \texttt{HELP}, EGR retrieves same-task entries and ranks them by
a normalized configuration distance:
\begin{equation}
\begin{aligned}
D(t,i)={}&
\frac{\|\mathbf{p}_t-\mathbf{p}_i\|_2}{s_p}
+w_R\frac{d_{\mathrm{SO}(3)}(\mathbf{R}_t,\mathbf{R}_i)}{s_R}\\
&+w_g\frac{\|\mathbf{g}_t-\mathbf{g}_i\|_2}{s_g}
+w_q\frac{\|\mathbf{q}_t-\mathbf{q}_i\|_2}{s_q},
\end{aligned}
\label{eq:reference_distance}
\end{equation}
where $\mathbf{p}$, $\mathbf{R}$, $\mathbf{g}$, and $\mathbf{q}$ denote
end-effector position, orientation, gripper width, and joint angles;
$d_{\mathrm{SO}(3)}$ is the rotation geodesic distance. The scales
$s_\cdot$ normalize heterogeneous quantities, and $w_\cdot$ weight
their contributions. Distance filtering and IK feasibility checks
produce up to five candidate reference states, each paired with its
external and wrist images. Configuration distance and IK filtering identify nearby,
kinematically feasible targets; Astra then assesses their suitability
for the observed task state.

\paragraph{Local Correction and Policy Resumption.}
Astra compares the candidate images with live views using the task
instruction, recent execution history, and a fresh VLA proposal.
Its visual assessment considers object arrangement, gripper--object
relations, and task stage when selecting a reference. For example,
live images can reveal a dropped object even when the arm remains
close to a successful reference pose. Astra selects one of the
three modes in Fig.~\ref{fig:reasoning_action_paradigms}d:
(A) \emph{no-op handback} when the VLA proposal is already suitable;
(B) \emph{reference-guided correction}, selecting a visually compatible
candidate and using bounded IK to approach its pose; or
(C) \emph{visual intervention}, generating end-effector delta
corrections from current images when the candidate set offers no
suitable target. This visual selection and fallback allow recovery
actions to adapt to the current scene.
Observations and VLA proposals are refreshed after each correction.
Reference-guided recovery returns control upon reaching the target
tolerance; visual intervention ends when Astra judges that the VLA can
resume progress. A progress safeguard requires physical correction
after repeated ineffective no-op handbacks or a forced request.
Task success is evaluated through subsequent closed-loop execution.

\subsection{Offline Learning}
\label{sec:offline}

The offline branch in Fig.~\ref{fig:method}a learns a failure-risk
monitor from recorded VLA rollouts and builds a recovery memory from
their successful states. Both are fixed during evaluation.

\paragraph{Failure-Risk Monitor.}
Each head projects scene-context, action-intent, and low-level features
into a fused representation, followed by a single-query GRU and scalar
risk output. We average independently trained heads:
\begin{equation}
p_t^{\mathrm{off}}=\frac{1}{K}\sum_{k=1}^{K}
\sigma\!\left(f_{\theta_k}(\mathbf{f}_t)\right).
\label{eq:offline_ensemble}
\end{equation}
A calibrated threshold converts this risk into an assistance vote,
which passes through temporal confirmation before a request is issued.

\paragraph{Trajectory-Balanced Supervision.}
For rollout $\tau_i$ with $T_i$ queries, the terminal label $y_i$ is
one for timeout failure and zero for success. All queries inherit
this weak outcome label. We normalize each trajectory's contribution
to prevent longer executions from dominating training:
\begin{equation}
\mathcal{L}_{\mathrm{off}}=
\frac{1}{N}\sum_{i=1}^{N}\frac{1}{T_i}
\sum_{t=1}^{T_i}\operatorname{BCE}\!\left(
\sigma(f_\theta(\mathbf{f}_{i,t})),y_i\right).
\label{eq:offline_objective}
\end{equation}
Terminal outcome labels provide weak supervision for assistance risk,
while stored successful states provide EGR references for recovery.
Calibration and temporal confirmation translate these predictions
into deployment decisions.
The CALL ablation retains this monitor but removes reference-guided
recovery; CALL+EGR uses both components.

\subsection{Online Self-Evolution}
\label{sec:online}

The online branch in Fig.~\ref{fig:method}b initializes a recurrent
request policy and an empty memory, learning assistance decisions and
recovery references from deployment experience while retaining the
same execution and recovery interface.

\paragraph{Recurrent Requests and Segment Feedback.}
A GRU encodes recent query features. Its actor additionally conditions
on causal context, including recent votes, cooldown, remaining VLA
budget, and intervention history; a critic estimates segment returns.
The actor maintains exploration through a bounded request probability:
\begin{equation}
\mu_\theta(b_t=1)=\epsilon_{\mathrm{exp}}+
(1-2\epsilon_{\mathrm{exp}})\sigma(z_t),
\label{eq:online_request}
\end{equation}
where $z_t$ is the actor logit.

A confirmed request ends the current VLA segment, and VLA resumption
starts a new one. Returns are computed within VLA-controlled segments,
with intervention boundaries separating successive segments.
Feedback rewards autonomous
success and requests followed by physical correction, penalizes VLA
timeouts, and assigns a delayed penalty when a no-op
handback is followed by autonomous success. Forced requests are
excluded from autonomous-request rewards. This feedback supplies
observed execution signals for the joint CALL update.

\paragraph{First-Alarm Marginalization.}
Temporal confirmation couples successive votes. We optimize an
outcome-based surrogate utility by marginalizing the first confirmed
request over voting histories, conditional on each recorded VLA
feature sequence $\tau$. Let $h$ encode
the finite confirmation history, $\mathcal{T}(h,b)$ update that
history, and $G_t(h,b)$ indicate a valid first alarm under the cooldown
constraints. Under behavior policy $\mu_{\mathrm{old}}$, unabsorbed
state occupancy evolves as
\begin{equation}
\begin{aligned}
d_{t+1}(h')=\sum_{h,b}&d_t(h)\mu_{\mathrm{old},t}(b\mid h)
[1-G_t(h,b)]\\
&\cdot\mathbb{I}\!\left[\mathcal{T}(h,b)=h'\right].
\end{aligned}
\label{eq:exact_forward}
\end{equation}
Probability mass is absorbed at the first alarm. Assigned segment
outcomes determine alarm and no-alarm utilities, encouraging requests
on failure-labeled segments and continued execution on success-labeled
segments. Backward dynamic programming computes the action-values and
advantages of this conditional surrogate:
\begin{equation}
\begin{aligned}
V_t(h)&=\sum_b\mu_{\mathrm{old},t}(b\mid h)Q_t(h,b),\\
A_t(h,b)&=Q_t(h,b)-V_t(h).
\end{aligned}
\label{eq:exact_advantage}
\end{equation}
We use an occupancy-weighted clipped update:
\begin{equation}
\begin{aligned}
\mathcal{L}_{\mathrm{FA}}=-\sum_{t,h,b}
&d_t^{\mathrm{old}}(h)\mu_{\mathrm{old},t}(b\mid h)\\
&\cdot\min\!\left(r_tA_t,\,
\operatorname{clip}(r_t,1-\epsilon,1+\epsilon)A_t\right),
\end{aligned}
\label{eq:exact_ppo}
\end{equation}
where $r_t=\mu_{\theta,t}(b\mid h)/\mu_{\mathrm{old},t}(b\mid h)$
and $A_t=A_t(h,b)$, with causal execution context recomputed for each
voting history conditional on the recorded feature sequence $\tau$.
Observed intervention feedback complements this outcome-based surrogate
in the joint update (Appendix~\ref{app:first_alarm_objective}).

\paragraph{Joint Policy and Memory Updates.}
We combine the first-alarm surrogate with an intervention-feedback PPO loss,
critic loss, and entropy bonus:
\begin{equation}
\begin{aligned}
\mathcal{L}_{\mathrm{online}}=\mathcal{L}_{\mathrm{FA}}
+\lambda_{\mathrm{fb}}\big(&\mathcal{L}_{\mathrm{PPO}}^{\mathrm{fb}}
+\lambda_v\mathcal{L}_{\mathrm{value}}\\
&-\lambda_h\mathcal{H}(\mu_\theta)\big).
\end{aligned}
\label{eq:online_objective}
\end{equation}
After each episode batch, eligible VLA segments update $\theta$, while
reference points from successful episodes expand $\mathcal{M}$:
\begin{equation}
\begin{aligned}
\theta_{k+1}&=\operatorname{Update}_{\mathrm{CALL}}
(\theta_k,\mathcal{D}_k^{\mathrm{upd}}),\\
\mathcal{M}_{k+1}&=\mathcal{M}_k\cup\mathcal{E}_k^{\mathrm{success}}.
\end{aligned}
\label{eq:self_evolution_updates}
\end{equation}
The reference set $\mathcal{E}_k^{\mathrm{success}}$ contains only
VLA-controlled states, excluding IK and Astra intervention segments.
The two updates refine request decisions and expand available recovery
references while keeping both foundation models frozen. Online
evaluation measures the resulting collaboration during deployment.

\section{Experiments}
\label{sec:experiments}

\subsection{Experimental Setup}
\label{sec:exp_setup}

\paragraph{Benchmarks and Baselines.}
We evaluate PICA in simulation on the LIBERO
benchmark~\cite{liu2023liberobenchmarkingknowledgetransfer} across all
four task suites (LIBERO-Spatial, LIBERO-Object, LIBERO-Goal, and
LIBERO-10) and in the real world on a physical SO-101 robot arm. Our
experiments test two core hypotheses: (1) selective reasoning and
experience-guided recovery improve manipulation reliability while
preserving high execution speed, and (2) online self-evolution can
learn effective assistance decisions and recovery priors from scratch
during deployment.

We compare the three baseline paradigms illustrated in
Fig.~\ref{fig:reasoning_action_paradigms}(a--c) against three
configurations of our framework (Fig.~\ref{fig:reasoning_action_paradigms}d
and Fig.~\ref{fig:method}):
\begin{itemize}
    \item \textbf{Frozen $\pi_{0.5}$} (Fig.~\ref{fig:reasoning_action_paradigms}a, Embodied Models):
    Standalone execution using the pretrained $\pi_{0.5}$ policy
    without external reasoning.
    \item \textbf{Astra Direct Control} (Fig.~\ref{fig:reasoning_action_paradigms}b, Frontier VLM):
    Closed-loop control driven entirely by GPT-6 Astra without the VLA.
    \item \textbf{Continuous Astra Supervision} (Fig.~\ref{fig:reasoning_action_paradigms}c, Call Every Step):
    Astra checks and approves or intervenes on every VLA action
    proposal before execution.
    \item \textbf{CALL}: Our selective collaboration framework using
    the offline-trained failure-risk monitor and visual Astra
    intervention without experience-guided recovery.
    \item \textbf{CALL+EGR} (Fig.~\ref{fig:method}a): Our full
    offline-trained framework combining the failure-risk monitor with
    experience-guided recovery (EGR) using the offline recovery memory.
    \item \textbf{Online Self-Evolution} (Fig.~\ref{fig:method}b): Our
    online learning configuration that initializes the recurrent
    request policy from scratch and builds the online recovery memory
    during execution.
\end{itemize}
The offline and online configurations are evaluated as independent
training regimes: CALL and CALL+EGR use pre-collected rollouts, whereas
Online Self-Evolution learns during deployment from a random CALL
initialization and an empty memory.

\paragraph{Temporal Confirmation Setting.}
Both regimes confirm an autonomous request when at least four of the
latest five votes are positive, subject to cooldown. Evaluation protocols and implementation details are provided in
Appendices~\ref{app:eval_protocol} and \ref{app:implementation}.

\paragraph{Evaluation Metrics.}
We measure task performance and computational efficiency using task
success rate (SR, \%), control frequency (Hz), API requests per episode
(API/Ep.), and total input-plus-output tokens per episode (Tokens). On
the physical robot, we also report the percentage of executed robot
actions controlled by Astra (Astra Actions, \%). To evaluate the CALL
policy independently of recovery, we report failure capture (recall on
failed trajectories), false-positive rate (FPR on successful
trajectories), precision, and episode-level average precision (AP).

\subsection{Efficient Reasoning--Action Collaboration}
\label{sec:exp_collaboration}

\paragraph{Overall Reliability and Efficiency.}
Table~\ref{tab:main_results} summarizes the results across the four
LIBERO suites. CALL+EGR achieves the highest overall success rate of
99.0\%, outperforming the frozen $\pi_{0.5}$ baseline by 3.5 percentage
points. The largest improvement occurs on the long-horizon LIBERO-10
suite, where success rises from 86.0\% to 98.0\%. CALL+EGR maintains
an execution frequency of 11.46 Hz while consuming 10.99k tokens and
0.47 API calls per episode.

\begin{table*}[t]
\centering
\caption{
Performance and efficiency comparison across the four LIBERO suites.
Success rates (SR) are in percentages. API/Ep.\ and Tokens (k) denote
the average API requests and total tokens (in thousands) per episode.
CALL uses selective intervention with visual recovery; CALL+EGR adds
experience-guided recovery. Astra Direct Control succeeds in 15 of
40 trials; its efficiency values are aggregated across all four suites.
}
\label{tab:main_results}
\resizebox{\textwidth}{!}{%
\begin{tabular}{lrrrrrrrr}
\toprule
\textbf{Method}
& \textbf{L-10}
& \textbf{Object}
& \textbf{Goal}
& \textbf{Spatial}
& \textbf{Overall SR $\uparrow$}
& \textbf{Hz $\uparrow$}
& \textbf{API/Ep.\ $\downarrow$}
& \textbf{Tokens (k) $\downarrow$} \\
\midrule
Frozen $\pi_{0.5}$
& 86.0 & 100.0 & 98.0 & 98.0
& 95.5 & 14.56 & 0.00 & 0.00 \\
Astra Direct Control
& 0.0 & 60.0 & 80.0 & 10.0
& 37.5 & 0.2138 & 63.08 & 12{,}218.0 \\
Continuous Astra Supervision
& 100.0 & 100.0 & 100.0 & 80.0
& 95.0 & 0.360 & 43.18 & 938.40 \\
CALL
& 98.0 & 100.0 & 100.0 & 96.0
& 98.5 & 6.82 & 1.05 & 19.26 \\
\textbf{CALL+EGR}
& 98.0 & 100.0 & 100.0 & 98.0
& \textbf{99.0} & 11.46 & 0.47 & 10.99 \\
\bottomrule
\end{tabular}%
}
\end{table*}

\paragraph{Benefit of Selective CALL Policy.}
Continuous Astra Supervision (Call Every Step) queries Astra before
each VLA action proposal executes, yielding 0.360 Hz and 938.40k tokens
per episode. Selective CALL achieves 98.5\% overall success compared
with 95.0\% for continuous supervision, while reducing token usage by
97.9\% and achieving $18.9\times$ its execution frequency
(6.82 Hz vs.\ 0.360 Hz).

\paragraph{Astra Direct Control.}
Astra Direct Control achieves 37.5\% success (15/40) in simulation at
0.2138 Hz, with 63.08 API calls and 12{,}218.0k tokens per episode.
In our evaluation, failures primarily occur because the fixed episode step
budget limits extended trial-and-error recovery, and long open-loop
action segments selected by Astra frequently disturb objects before the
next observation is acquired. Recent empirical studies on GPT-6
Astra~\cite{galbotteam2026systematicallyexploringcapabilitiesgpt6,yang2026astranomy}
similarly observe that higher direct-control success relies on
substantially expanded step budgets (up to $10\times$ the standard
horizon) and finer-grained replanning, which correspondingly increases
token consumption and inference latency.

\paragraph{Benefit of Experience-Guided Recovery.}
Comparing CALL+EGR with CALL in Table~\ref{tab:main_results} highlights
the impact of experience-guided recovery. Adding EGR reduces token
consumption by 42.9\% (from 19.26k to 10.99k), cuts API requests by
55.2\% (from 1.05 to 0.47), and increases execution frequency to
$1.68\times$ that of CALL (from 6.82 Hz to 11.46 Hz), while raising overall success
to 99.0\%. Across all evaluated episodes, EGR reduces the total number
of direct visual correction steps from 2,209 to 180. By pairing
selective invocation with configuration-based reference retrieval, EGR
resolves most interventions through bounded IK corrections toward a
visually compatible state rather than multi-step visual replanning.

\subsection{Online Self-Evolution}
\label{sec:exp_self_evolution}

\paragraph{Learning Collaboration from Scratch.}
Separately, we evaluate Online Self-Evolution (Fig.~\ref{fig:method}b) on
LIBERO-10. As the robot executes tasks, outcome and intervention
feedback from eligible VLA segments update the request policy every 10
episodes, while $\pi_{0.5}$ reference points from successful episodes
are appended to the online recovery memory.

\paragraph{Performance on LIBERO-10.}
Table~\ref{tab:online_results} compares Online Self-Evolution over 200
continual episodes against the fixed baselines on LIBERO-10. Online
Self-Evolution achieves a 94.0\% success rate, outperforming the frozen
$\pi_{0.5}$ policy by 8.0 percentage points while operating at 6.95 Hz
with 0.43 API calls and 6.09k tokens per episode. Note that this overall
rate averages the entire 200-episode learning rollout, including the
initial cold-start episodes.

\begin{table}[t]
\centering
\caption{
Comparison of online self-evolution and fixed methods on LIBERO-10.
Online Self-Evolution starts with a randomly initialized CALL policy
and an empty recovery memory. Astra Direct Control is evaluated on
ten LIBERO-10 trials (0/10 successes).
Table~\ref{tab:main_results} reports its four-suite aggregates.
}
\label{tab:online_results}
\resizebox{\columnwidth}{!}{%
\begin{tabular}{lrrrr}
\toprule
\textbf{Method}
& \textbf{SR (\%)}
& \textbf{Hz}
& \textbf{API/Ep.}
& \textbf{Tokens (k)} \\
\midrule
Frozen $\pi_{0.5}$
& 86.0 & 14.74 & 0.00 & 0.00 \\
Astra Direct Control
& 0.0 & 0.2674 & 75.40 & 358.64 \\
Continuous Astra Supervision
& 100.0 & 0.373 & 52.30 & 1150.90 \\
CALL
& 98.0 & 5.80 & 2.34 & 48.83 \\
CALL+EGR
& 98.0 & 9.80 & 1.24 & 30.09 \\
\midrule
\textbf{Online Self-Evolution}
& 94.0 & 6.95 & 0.43 & 6.09 \\
\bottomrule
\end{tabular}%
}
\end{table}

\paragraph{Rollout Progression.}
Table~\ref{tab:online_evolution} tracks task success and intervention
counts across intervals of the 200-episode online rollout. In the early
episodes (1--50), the uncalibrated request policy triggers few
autonomous \texttt{HELP} calls, relying more on forced requests when
the VLA budget is exhausted. Later intervals show increased use of
autonomous requests: in episodes 151--200, autonomous CALLs rise to 20
while forced requests drop to 4, and the interval success rate reaches
98.0\%.

\begin{table}[t]
\centering
\caption{
Progression of task success and assistance events across online
training intervals on LIBERO-10.
}
\label{tab:online_evolution}
\small
\setlength{\tabcolsep}{4pt}
\resizebox{\columnwidth}{!}{%
\begin{tabular}{lrrrr}
\toprule
\textbf{Episodes}
& \textbf{Success}
& \textbf{SR (\%)}
& \textbf{\shortstack{Autonomous\\CALL Count}}
& \textbf{\shortstack{Forced\\Assistance Count}} \\
\midrule
1--20    & 18/20   & 90.0 & 1  & 3 \\
21--50   & 26/30   & 86.7 & 3  & 8 \\
51--100  & 47/50   & 94.0 & 12 & 5 \\
101--150 & 48/50   & 96.0 & 8  & 8 \\
151--200 & 49/50   & 98.0 & 20 & 4 \\
\midrule
Total    & 188/200 & 94.0 & 44 & 28 \\
\bottomrule
\end{tabular}%
}
\end{table}

\paragraph{Checkpoint Evaluation on Shared Trajectories.}
To isolate the learning progress of the recurrent request policy from
changing rollout states, we evaluate frozen online checkpoints (after
20, 50, 100, 150, and 200 training episodes) on a fixed set of 100
uninterrupted $\pi_{0.5}$ trajectories on LIBERO-10. Conditional on
these fixed trajectories, we compute expected detection metrics by
marginalizing over the first-confirmed-request voting histories of
the stochastic CALL policy.

As shown in Table~\ref{tab:online_checkpoint_detection}, failure
capture improves steadily from 10.15\% at episode 20 to 97.39\% at
episode 200, while precision rises from 18.94\% to 60.88\%. Between
episodes 100 and 200, the CALL policy simultaneously increases failure
capture (from 63.58\% to 97.39\%) and reduces false alarms (FPR drops
from 22.38\% to 6.95\%, and expected alarmed episodes fall from 26.50
to 16.00). Between episodes 150 and 200, the total number of alarmed
episodes stays nearly constant ($\approx 16$), yet capture and
precision continue to improve, indicating sharper discrimination between
failing and solvable trajectories.
Appendix~\ref{sec:exp_call} reports additional architecture, feature, and
fixed-data replay ablations.

\begin{table}[t]
\centering
\caption{
Offline evaluation of online CALL checkpoints on 100 shared pure-VLA
trajectories. Exp.\ Alarmed Ep.\ is the expected number of trajectories
triggering an autonomous alarm. Capture, FPR, and Precision are in
percentages.
}
\label{tab:online_checkpoint_detection}
\small
\setlength{\tabcolsep}{4pt}
\resizebox{\columnwidth}{!}{%
\begin{tabular}{rrrrr}
\toprule
\textbf{\shortstack{Training\\Episodes}}
& \textbf{\shortstack{Exp.\ Alarmed\\Ep.}}
& \textbf{Capture $\uparrow$}
& \textbf{FPR $\downarrow$}
& \textbf{Precision $\uparrow$} \\
\midrule
20  & 5.36  & 10.15 & 4.83  & 18.94 \\
50  & 7.13  & 16.11 & 6.14  & 22.59 \\
100 & 26.50 & 63.58 & 22.38 & 23.99 \\
150 & 15.93 & 90.65 & 7.63  & 56.90 \\
200 & 16.00 & 97.39 & 6.95  & 60.88 \\
\bottomrule
\end{tabular}%
}
\end{table}

\subsection{Real-World Robot Experiments}
\label{sec:exp_real_world}

\begin{figure}[t]
    \centering
    \includegraphics[width=\columnwidth]{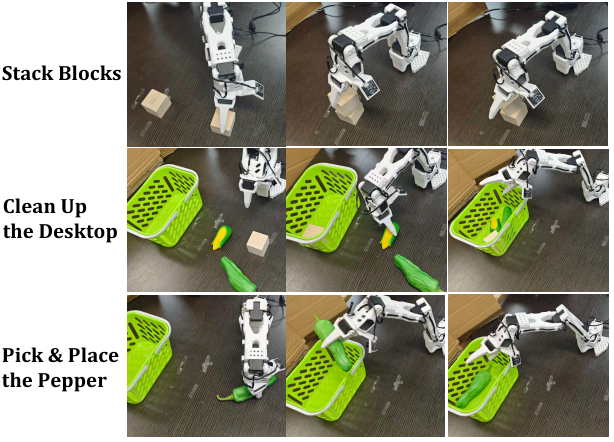}
    \caption{
        \textbf{Real-world manipulation tasks on the SO-101 robot.}
        From top to bottom: block stacking, desktop cleanup, and pepper
        pick-and-place, showing representative execution stages.
    }
    \label{fig:real_world_tasks}
\end{figure}

\paragraph{Experimental Setup.}
We evaluate PICA on a physical SO-101 robot arm across three tabletop
manipulation tasks: block stacking, desktop cleanup, and pepper
pick-and-place (Fig.~\ref{fig:real_world_tasks}). We run 30 trials per
method across four configurations: Frozen $\pi_{0.5}$, Astra Direct
Control, CALL+EGR, and Online Self-Evolution. In addition to success
rate (SR), control frequency (Hz), and token usage (in millions), we
report the percentage of executed robot steps controlled by Astra
(Astra Actions).

\begin{table}[t]
\centering
\caption{
Real-world results on the SO-101 robot arm over 30 trials per method.
Tokens (M) denotes average token consumption per episode in millions.
Astra Actions (\%) measures the fraction of executed steps controlled
by Astra.
}
\label{tab:real_world_results}
\resizebox{\columnwidth}{!}{%
\begin{tabular}{lrrrr}
\toprule
\textbf{Method}
& \textbf{SR (\%)}
& \textbf{Hz $\uparrow$}
& \textbf{Tokens (M) $\downarrow$}
& \textbf{\shortstack{Astra\\Actions (\%) $\downarrow$}} \\
\midrule
Frozen $\pi_{0.5}$
& 70.0 & \textbf{14.72} & 0.00 & 0.0 \\
Astra Direct Control
& 86.7 & 0.35 & 4.57 & 100.0 \\
\textbf{CALL+EGR}
& \textbf{93.3} & 12.60 & \textbf{0.43} & \textbf{13.6} \\
Online Self-Evolution
& 86.7 & 8.41 & 0.69 & 19.7 \\
\bottomrule
\end{tabular}%
}
\end{table}

\paragraph{Real-World Reliability and Efficiency.}
As shown in Table~\ref{tab:real_world_results}, CALL+EGR achieves the
highest real-world success rate of 93.3\% (28/30), improving over
Frozen $\pi_{0.5}$ (70.0\%) by 23.3 percentage points and Astra Direct
Control (86.7\%) by 6.6 percentage points. Compared with Astra Direct
Control, CALL+EGR achieves $36.0\times$ its execution frequency
(12.60 Hz vs.\ 0.35 Hz) and
reduces token usage by 90.6\% (0.43M vs.\ 4.57M). Only 13.6\% of
executed actions are controlled by Astra, with the VLA handling the
majority of physical execution.

\paragraph{Real-World Online Self-Evolution.}
Online Self-Evolution achieves 86.7\% success (26/30) on the physical
robot, outperforming Frozen $\pi_{0.5}$ by 16.7 percentage points. It
runs at 8.41 Hz with 19.7\% Astra-controlled actions, matching the
success rate of Astra Direct Control while cutting token consumption by
84.9\% and achieving $24.0\times$ its execution frequency.

\section{Conclusion}
\label{sec:conclusion}

We presented PICA, a selective reasoning--action collaboration
framework that pairs the fast visuomotor control of $\pi_{0.5}$ with
the reasoning and recovery capabilities of GPT-6 Astra. PICA uses a lightweight,
policy-aware CALL policy to request reasoning based on estimated
failure risk, complemented by a timeout safeguard. Experience-guided recovery (EGR) combines configuration-based
retrieval with visual selection of reference-guided IK or direct
correction to support early VLA resumption. Alongside offline
learning, Online Self-Evolution
allows the CALL policy and recovery memory to improve from scratch
during deployment without pre-collected failure data. Experiments across
four LIBERO suites and a physical SO-101 robot show that PICA
substantially improves manipulation success while keeping inference cost
low and execution speed high.

{
    \small
    \bibliographystyle{ieeenat_fullname}
    \bibliography{main}
}

\clearpage

\appendix

\setcounter{table}{0}
\renewcommand{\thetable}{A\arabic{table}}
\providecommand{\theHtable}{}
\renewcommand{\theHtable}{appendix.\arabic{table}}
\setcounter{equation}{0}
\renewcommand{\theequation}{A\arabic{equation}}
\renewcommand{\theHequation}{appendix.\arabic{equation}}


\section{Evaluation Protocol}
\label{app:eval_protocol}

\subsection{Simulation Evaluation Details}
\label{app:collaboration_protocol}

\paragraph{Evaluation Episodes.}
In the multi-suite evaluation (Table~\ref{tab:main_results}), Frozen
$\pi_{0.5}$, CALL, and CALL+EGR are evaluated over 5 episodes per task
across all four LIBERO suites (40 tasks total), yielding 200 episodes
per method. Due to the high inference cost of continuous reasoning queries,
Astra Direct Control and Continuous Astra Supervision are evaluated for
1 episode per task across the four suites (40 episodes per method). For
the LIBERO-10 comparison (Table~\ref{tab:online_results}), Frozen $\pi_{0.5}$, CALL, and
CALL+EGR use 50 episodes (5 per task), Astra Direct Control and
Continuous Astra Supervision use 10 episodes (1 per task), and Online
Self-Evolution is evaluated across 200 continual episodes (20 per
task).

\paragraph{Episode Step Budgets.}
Simulation evaluation uses a maximum episode horizon of 1,300
environment steps. In VLA--Astra collaboration, the forced-assistance
threshold for $\pi_{0.5}$ is 520 VLA execution steps. Reaching this
threshold triggers assistance, subject to the overall episode limit.
The total episode budget includes VLA execution, recovery, and policy
resumption.

\paragraph{Astra Inference and Adaptive Action Horizon.}
The simulation Astra Direct Control baseline uses GPT-6 Astra with
\texttt{xhigh} reasoning effort. Simulation configurations that combine
Astra with the VLA use \texttt{medium} reasoning effort. Astra
autonomously selects an execution length between 5 and 50 environment
steps according to the current task stage, allowing longer execution
segments for coarse reaching and shorter segments for contact-rich
manipulation or precision recovery.

\paragraph{Ablation Configurations (CALL vs.\ CALL+EGR).}
Both CALL and CALL+EGR use the same offline-trained failure-risk
monitor ensemble (Fig.~\ref{fig:method}a). When a \texttt{HELP}
intervention is triggered, CALL relies solely on Astra's visual
end-effector delta corrections (Mode C in Fig.~\ref{fig:reasoning_action_paradigms}d) without access to
the offline recovery memory $\mathcal{M}$ or reference-guided inverse
kinematics (IK). CALL+EGR enables the full experience-guided recovery
module (Mode B), including same-task reference retrieval from
$\mathcal{M}$ and bounded IK planning toward the selected reference
state. Both configurations allow no-op handback (Mode A) when Astra
determines no correction is required.

\paragraph{Online Training Protocol.}
Online Self-Evolution (Fig.~\ref{fig:method}b) starts with a
randomly initialized recurrent request policy and an empty online
recovery memory $\mathcal{M}=\emptyset$. During the 200-episode rollout
on LIBERO-10, the request policy is updated every 10 episodes (20
update rounds and 80 optimizer steps in total) using eligible VLA
segments, while $\pi_{0.5}$ reference points from newly completed
successful episodes (excluding IK and Astra segments) are appended to
$\mathcal{M}$.

\paragraph{Metric Definitions.}
Let $N$ be the number of evaluated episodes, $A$ the total executed
environment steps, $T$ the total wall-clock execution time, and
$K_{\mathrm{in}}, K_{\mathrm{out}}$ the total input and output tokens
across all $N$ episodes. We compute control frequency as $\mathrm{Hz} =
A/T$, token consumption (in thousands per episode) as $(K_{\mathrm{in}}
+ K_{\mathrm{out}}) / (1000 N)$, and API requests per episode
($\mathrm{API/Ep.}$) as the total number of Astra API calls divided by
$N$ (where a single CALL intervention may span multiple API calls).
Overall SR is the mean success rate across the four LIBERO suites.


\subsection{Real-World Evaluation Details}
\label{app:real_world_protocol}

\paragraph{Hardware Platform and Tasks.}
We evaluate PICA on a physical SO-101 robot arm across three tabletop
manipulation tasks: block stacking, desktop cleanup, and pepper
pick-and-place. Each method (Frozen $\pi_{0.5}$, Astra Direct Control,
CALL+EGR, and Online Self-Evolution) is evaluated over 30 trials (10
trials per task).

\paragraph{Offline Data Collection and Memory Initialization.}
To train the offline failure-risk monitor and initialize the offline
recovery memory for CALL+EGR on the real robot, we collect 60 rollouts
using the frozen $\pi_{0.5}$ policy (40 successful and 20 failed
trajectories). The offline failure-risk monitor is trained on all 60
trajectories using terminal outcome labels, while the offline recovery
memory $\mathcal{M}$ is populated using intermediate non-terminal robot
states and camera images from the 40 successful trajectories. Online
Self-Evolution learns its recurrent request policy and populates
$\mathcal{M}$ during the 30 online trials.

\paragraph{Execution Budgets and Metrics.}
Real-world evaluation on the SO-101 robot uses a maximum episode
horizon of 2,000 control steps. In VLA--Astra collaboration, the
forced-assistance threshold for $\pi_{0.5}$ is 650 VLA execution
steps. All real-world configurations involving Astra use
\texttt{xhigh} reasoning effort. Astra autonomously selects an
execution length between 5 and 50 control steps according to the
current task stage. In addition to SR, Hz, and total tokens per episode
(in millions), we report Astra Actions (\%), defined as the number of
control steps executed by Astra divided by the total executed steps
across the episode.


\subsection{Online Checkpoint Evaluation Protocol}
\label{app:checkpoint_protocol}

\paragraph{Shared Pure-VLA Trajectory Set.}
To measure how online training improves failure detection independently
of Astra's recovery actions, we evaluate frozen online CALL checkpoints
(saved after 20, 50, 100, 150, and 200 online episodes) on a shared set
of 100 uninterrupted $\pi_{0.5}$ rollouts on LIBERO-10 (10 rollouts per
task, up to 520 steps each, comprising 10 failed and 90 successful
trajectories). These trajectories record uninterrupted VLA execution until success
or the 520-step limit.

\paragraph{Expected Detection Metrics.}
Because the online recurrent request policy outputs a stochastic vote
probability at each query step, we evaluate each checkpoint by
propagating probability mass across all $2^4=16$ possible voting
histories along each trajectory, absorbing mass when the 4-of-5 voting
rule is first satisfied.

Let $p_i\in[0,1]$ be the resulting probability that trajectory $i$
triggers at least one autonomous CALL alarm, and let $\mathcal{F}$
($|\mathcal{F}|=10$) and $\mathcal{S}$ ($|\mathcal{S}|=90$) denote the
sets of failed and successful trajectories. The expected number of
alarmed episodes ($\bar{N}_{\mathrm{alarm}}$), expected true positives
($\overline{\mathrm{TP}}$), and expected false positives
($\overline{\mathrm{FP}}$) are given by
\begin{equation}
\begin{aligned}
\bar{N}_{\mathrm{alarm}} &= \sum_{i=1}^{100}p_i, \\
\overline{\mathrm{TP}} &= \sum_{i\in\mathcal{F}}p_i, \qquad
\overline{\mathrm{FP}} = \sum_{i\in\mathcal{S}}p_i.
\end{aligned}
\label{eq:expected_alarm_counts}
\end{equation}
From these expected counts, we compute failure capture as
$\overline{\mathrm{TP}}/|\mathcal{F}|$, false-positive rate (FPR) as
$\overline{\mathrm{FP}}/|\mathcal{S}|$, and precision as
$\overline{\mathrm{TP}}/\bar{N}_{\mathrm{alarm}}$. Precision is defined
when $\bar{N}_{\mathrm{alarm}}>0$.


\section{Additional CALL Policy Ablations}
\label{sec:exp_call}

\subsection{Offline Architecture and Feature Ablations}

We evaluate offline failure-risk monitor architectures and input
feature combinations on 800 held-out pure-VLA trajectories across all
four LIBERO suites (779 successful and 21 failed trajectories). All
variants use the 4-of-5 voting rule, with decision thresholds
calibrated on a separate validation split to target a 5\% FPR on
successful episodes. As shown in the bottom row of
Table~\ref{tab:offline_call_ablation}, the offline failure-risk monitor
introduced in Sec.~\ref{sec:offline}
detects 20 out of 21 failures (95.24\% capture) at a 4.49\% FPR,
achieving 36.36\% precision and 71.96\% episode-level average precision
(AP).

\paragraph{Backbone Architecture and History Length.}
The top block of Table~\ref{tab:offline_call_ablation} reports the mean
and standard deviation across three random seeds for an MLP baseline
and GRUs with sequence lengths $L\in\{1,4,8,16,32\}$. The single-query
GRU ($L=1$, which relies on causal transition features
$\boldsymbol{\delta}_t$ for short-term temporal context) achieves the
same 93.65\% capture as the MLP while lowering FPR to 4.36\% and
improving precision to 37.33\%. Both $L=16$ and $L=32$ reach 96.83\%
capture, with AP values of 79.51\% and 79.66\% and FPR values of 5.61\%
and 5.73\%, respectively. CPU inference latency increases from 0.40 ms
at $L=1$ to 2.06 ms at $L=32$. We use $L=1$ for the lightweight offline
monitor ensemble and $L=32$ to provide longer temporal context for the
online recurrent request policy.

\paragraph{Input Feature Modalities.}
The middle block of Table~\ref{tab:offline_call_ablation} ablates the
components of the 3165-dimensional policy query feature vector
$\mathbf{f}_t$ using the $L=1$ GRU. Removing the action-intent features
($\mathbf{h}_t^{\mathrm{act}}$) causes the largest drop in failure
capture (from 93.65\% to 88.89\%), indicating that internal flow-matching
action tokens provide useful failure cues. Conversely, using action
features alone or removing scene-context (semantic) features
($\mathbf{h}_t^{\mathrm{sem}}$) achieves 95.24\% capture and higher AP,
but increases false alarms (FPR rises from 4.36\% to 5.73\% and
5.13\%, respectively) at the calibrated operating threshold. Similarly,
removing causal transition features ($\boldsymbol{\delta}_t$) increases
FPR from 4.36\% to 5.13\%. The full representation combines high failure capture with a
relatively low false-positive rate at the selected operating point.

\begin{table*}[t]
\centering
\caption{
Offline CALL policy ablations over 800 held-out pure-VLA trajectories
across all four LIBERO suites. All metrics are in percentages. Ablation
rows report mean $\pm$ standard deviation over three training seeds;
the bottom row reports the three-head ensemble ($L=1$).
AP denotes episode-level average precision.
}
\label{tab:offline_call_ablation}
\small
\setlength{\tabcolsep}{8pt}
\resizebox{\textwidth}{!}{%
\begin{tabular}{lcccc}
\toprule
\textbf{Configuration}
& \textbf{Capture $\uparrow$}
& \textbf{FPR $\downarrow$}
& \textbf{Precision $\uparrow$}
& \textbf{AP $\uparrow$} \\
\midrule

\multicolumn{5}{l}{
\textit{Backbone Architecture and History Length}
} \\

MLP
& $93.65\pm2.75$ & $4.54\pm1.22$
& $36.45\pm5.73$ & $73.84\pm1.86$ \\

GRU ($L=1$)
& $93.65\pm2.75$ & $4.36\pm1.12$
& $37.33\pm5.96$ & $70.62\pm1.72$ \\

GRU ($L=4$)
& $95.24\pm0.00$ & $5.22\pm1.16$
& $33.49\pm5.22$ & $76.02\pm1.40$ \\

GRU ($L=8$)
& $95.24\pm0.00$ & $4.79\pm0.85$
& $35.21\pm4.28$ & $78.00\pm2.39$ \\

GRU ($L=16$)
& $96.83\pm2.75$ & $5.61\pm0.91$
& $32.02\pm3.25$ & $79.51\pm2.78$ \\

GRU ($L=32$)
& $96.83\pm2.75$ & $5.73\pm0.85$
& $31.47\pm2.58$ & $79.66\pm2.05$ \\

\midrule

\multicolumn{5}{l}{
\textit{Input Feature Ablation (GRU, $L=1$)}
} \\

Full representation ($\mathbf{f}_t$)
& $93.65\pm2.75$ & $4.36\pm1.12$
& $37.33\pm5.96$ & $70.62\pm1.72$ \\

Scene context / Semantic only ($\mathbf{h}_t^{\mathrm{sem}}$)
& $90.48\pm0.00$ & $5.43\pm1.07$
& $31.39\pm4.54$ & $63.76\pm5.53$ \\

Action intent only ($\mathbf{h}_t^{\mathrm{act}}$)
& $95.24\pm0.00$ & $5.73\pm0.98$
& $31.21\pm3.57$ & $79.36\pm2.32$ \\

Semantic + Action
& $92.06\pm5.50$ & $4.88\pm1.00$
& $34.10\pm3.64$ & $70.00\pm1.28$ \\

Without Semantic
& $95.24\pm0.00$ & $5.13\pm0.89$
& $33.65\pm4.16$ & $80.88\pm1.88$ \\

Without Action
& $88.89\pm2.75$ & $4.28\pm0.73$
& $36.21\pm4.67$ & $66.98\pm3.70$ \\

Without Proprioception/Statistics ($\mathbf{s}_t, \mathbf{u}_t$)
& $92.06\pm5.50$ & $4.19\pm1.09$
& $37.79\pm5.04$ & $69.53\pm1.73$ \\

Without Causal Transitions ($\boldsymbol{\delta}_t$)
& $93.65\pm2.75$ & $5.13\pm0.59$
& $33.07\pm1.98$ & $68.80\pm2.48$ \\

\midrule

\textbf{Offline Failure-Risk Monitor (Ours)}
& 95.24 & 4.49 & 36.36 & 71.96 \\

\bottomrule
\end{tabular}%
}
\end{table*}


\subsection{First-Alarm Objective Ablation via Offline Replay}

To examine the first-confirmed-request surrogate objective
(Eq.~\ref{eq:exact_ppo}) under fixed-data replay, we compare it against
step-wise request-level supervision on a fixed replay dataset of 200
online collaboration episodes, and evaluate the resulting CALL policies
on 1,200 pure-VLA LIBERO-10 trajectories (including 110 failures). We
test both objectives under two buffer strategies: updating on recent
episodes only (\textit{Recent}) versus replaying all accumulated
episodes (\textit{Full History}). Decision thresholds are calibrated on
a separate validation split to target a 5\% FPR.

As shown in Table~\ref{tab:online_call_ablation}, first-alarm
optimization achieves higher failure capture under both buffer
strategies: capture increases from 56.06\% to 83.03\% with recent-episode
updates, and from 26.97\% to 80.91\% with full-history replay. The
corresponding FPR increases from 3.12\% to 4.28\% and from 4.46\% to
6.12\%, respectively; precision also improves under both strategies.
Dynamic programming marginalizes first-confirmed-request voting
histories for fixed recorded VLA features and assigned utilities,
incorporating the deployed confirmation rule into the surrogate.
Among the first-alarm variants, recent-episode updates achieve higher
capture (83.03\% vs.\ 80.91\%) and lower FPR (4.28\% vs.\ 6.12\%) than
full-history replay.

\begin{table}[t]
\centering
\caption{
Comparison of step-wise request-level supervision and first-alarm
marginalization under fixed-data replay (evaluated on 1,200 pure-VLA
LIBERO-10 trajectories). Capture and FPR report mean $\pm$ standard
deviation over three seeds. All metrics are in percentages.
}
\label{tab:online_call_ablation}
\resizebox{\columnwidth}{!}{%
\begin{tabular}{lccc}
\toprule
\textbf{Optimization Objective}
& \textbf{Capture $\uparrow$}
& \textbf{FPR $\downarrow$}
& \textbf{Precision $\uparrow$} \\
\midrule

Request-level (Recent)
& $56.06\pm11.86$
& $3.12\pm0.78$
& $64.22$ \\

Request-level (Full History)
& $26.97\pm6.70$
& $4.46\pm0.68$
& $37.60$ \\

\textbf{First-Alarm (Recent)}
& $\mathbf{83.03\pm5.01}$
& $4.28\pm0.78$
& $\mathbf{66.37}$ \\

First-Alarm (Full History)
& $80.91\pm0.91$
& $6.12\pm0.83$
& $57.30$ \\

\bottomrule
\end{tabular}%
}
\end{table}


\subsection{Pure PPO Comparison via Offline Replay}
\label{app:pure_ppo_replay}

We compare the first-alarm objective with Pure PPO under fixed-data
replay. Both methods use the same recurrent CALL architecture,
confirmation rule and cooldown, alarm utilities, and sequence of
200 recorded collaboration episodes. Both update on the latest batch
of ten episodes, use only VLA segments ending in observed autonomous
success or timeout, and take 80 Adam steps per seed. Pure PPO samples
votes from the current policy until the first confirmed request or
segment termination. It optimizes a clipped policy objective with a
value baseline and entropy regularization. First-Alarm (Recent)
optimizes the actor by marginalizing over the admissible vote histories.

We evaluate the final checkpoints from three random seeds on the 1,200
pure-VLA LIBERO-10 trajectories in Table~\ref{tab:online_call_ablation}
(110 failures). Thresholds are calibrated on 400 separate trajectories
to target a 5\% false-positive rate on successful episodes. At test time,
thresholded votes are passed through the 4-of-5 confirmation rule.

\begin{table}[t]
\centering
\caption{Pure PPO and First-Alarm (Recent) under fixed-data replay.
Values are means over three seeds. Capture and FPR also report sample
standard deviations. CALL count is the number of trajectories with
at least one autonomous request out of 1,200. Rates and precision are
percentages.}
\label{tab:pica_sampled_ppo_replay}
\resizebox{\columnwidth}{!}{%
\begin{tabular}{lrrrrr}
\toprule
\textbf{Method}
& \textbf{CALL Count}
& \textbf{CALL Rate}
& \textbf{Capture $\uparrow$}
& \textbf{FPR $\downarrow$}
& \textbf{Precision $\uparrow$} \\
\midrule
Pure PPO
& 133.33 & 11.11 & $53.64\pm28.18$ & $6.82\pm1.65$ & 42.59 \\
First-Alarm (Recent)
& 138.00 & 11.50 & $83.03\pm5.01$ & $4.28\pm0.78$ & 66.37 \\
\bottomrule
\end{tabular}%
}
\end{table}

At similar CALL rates (11.50\% vs.\ 11.11\%),
First-Alarm (Recent) improves failure capture over Pure PPO by
29.39 percentage points (paired 95\% CI: [5.24, 50.25]) and
increases mean precision from 42.59\% to 66.37\%
(Table~\ref{tab:pica_sampled_ppo_replay}). We estimate the capture
interval by resampling tasks and source groups jointly across methods
with the three training seeds held fixed.

\section{Implementation and Evaluation Details}
\label{app:implementation}

\subsection{Astra Direct Control in Simulation}
\label{app:astra_direct}

The Astra Direct Control baseline in simulation uses GPT-6 Astra
with reasoning effort set to \texttt{xhigh}.
Astra selects actions and their execution lengths without a VLA.
Evaluation comprises 40 trials, with ten trials in each LIBERO suite.
It succeeds in 0/10 LIBERO-10, 6/10 Object, 8/10 Goal, and 1/10 Spatial
trials, yielding 15/40 successes (37.5\%) overall.
The reported four-suite aggregates are 0.2138 Hz, 63.08 API requests
per episode, and 12{,}218.0k tokens per episode.

On LIBERO-10, Astra Direct Control achieves 0/10 successes (0.0\%),
with a control frequency of 0.2674 Hz, 75.40 API requests per episode,
and 358.64k tokens per episode (Table~\ref{tab:online_results}).
The aggregate efficiency values in Table~\ref{tab:main_results}
summarize evaluation across all four suites.

\subsection{CALL Representation and Temporal Confirmation}
\label{app:call_settings}

The feature blocks in Eq.~\ref{eq:call_representation} have dimensions
2048 (scene context), 1024 (action intent), 8 (proprioception),
36 (action-chunk statistics), and 49 (causal transitions), yielding
3165 features per query. The VLA predicts a ten-step action chunk and
replans every five environment steps. CALL evaluates the query
features before the corresponding five-step action prefix executes.

Both offline and online configurations use four positive votes in
the most recent five queries to confirm an autonomous request, with
a post-intervention cooldown. Forced requests are triggered separately
when the VLA budget is exhausted. For first-confirmed-request marginalization,
the confirmation history consists of the preceding four binary votes,
giving at most 16 history states. The transition operator shifts this
history and appends the current vote; probability mass is absorbed
when the confirmation and cooldown conditions permit an alarm.

\subsection{Offline Training and Recovery Settings}
\label{app:offline_settings}

The offline monitor projects scene-context, action-intent, and
low-level features into 96, 96, and 64 dimensions, respectively.
Their concatenation is mapped to a 192-dimensional hidden state and
processed by a single-query GRU with a scalar risk head.
Each head contains 573,889 trainable parameters. We independently train
three heads ($K=3$) on the same 600 recorded VLA rollouts across 40
LIBERO tasks using AdamW with a learning rate of $3\times10^{-4}$.
At each query, we average the three sigmoid probabilities and apply a
calibrated risk threshold to obtain one vote for temporal confirmation.

For Eq.~\ref{eq:reference_distance}, we use
$s_p=0.05$, $s_R=0.35$, $s_g=0.02$,
$s_q=0.35\sqrt{7}$, and $w_R=w_g=w_q=0.25$.
Reference configurations contain seven arm joint angles,
end-effector pose, gripper width, and optional joint velocities.
Candidates with $D(t,i)>6$ are discarded. We check IK feasibility for
the top 15 remaining candidates and provide up to five feasible
references to Astra, together with their external and wrist images.
Astra compares these candidates with live images and the task
instruction to select an IK target consistent with the current
object configuration and manipulation stage. A selected reference
provides a pose target for bounded IK correction. When all candidates
are unsuitable, Astra generates direct visual corrections from the
current observations. A suitable VLA proposal permits immediate
handback.

\subsection{Online Training Settings}
\label{app:online_settings}

The online policy processes histories of up to 32 query features.
Scene-context, action-intent, and low-level features are projected
into 192, 192, and 128 dimensions, fused into a 384-dimensional
representation, and processed by a GRU. The actor also receives
20 causal context variables encoding recent votes, cooldown,
remaining VLA budget, and intervention history. We set
$\epsilon_{\mathrm{exp}}=0.05$ in Eq.~\ref{eq:online_request}.

The policy is updated every ten episodes using eligible VLA segments
from the latest batch. Segment feedback assigns $+0.2$ to autonomous
VLA success, $-4$ to VLA budget timeout, and $+1$ to an autonomous
request followed by physical correction. A no-op handback receives a
delayed penalty of $-1$ if the resumed VLA subsequently succeeds
autonomously. Forced requests are excluded from autonomous-request
rewards. For the first-alarm surrogate, failure-labeled segments have
alarm/no-alarm utilities of $+1/-4$, whereas success-labeled segments have
utilities of $-1/+0.2$.

The online recovery memory starts empty. After each batch, reference
points from VLA-controlled portions of successful episodes are
appended; IK and Astra intervention segments are excluded.

\subsection{Conditional First-Alarm Objective}
\label{app:first_alarm_objective}

Let $\tau=(\mathbf{f}_1,\ldots,\mathbf{f}_T)$ be a recorded VLA
feature sequence with assigned outcome label $y$, and let
$T_{\mathrm{alarm}}$ denote the time of the first confirmed request.
Writing $u_{\mathrm{alarm}}(y)$ and $u_{\mathrm{none}}(y)$ for
the utilities above, the conditional surrogate is
\begin{equation}
\begin{aligned}
J_{\mathrm{FA}}(\theta\mid\tau,y)
={}&\sum_{t=1}^{T}
P_\theta(T_{\mathrm{alarm}}=t\mid\tau)
u_{\mathrm{alarm}}(y)\\
&+P_\theta(T_{\mathrm{alarm}}>T\mid\tau)
u_{\mathrm{none}}(y).
\end{aligned}
\label{eq:first_alarm_surrogate}
\end{equation}
The probabilities also condition on the segment's initial confirmation
state and execution context. Forward propagation sums the admissible
voting histories; backward recursion computes values for the
specified utilities.

For a given outcome label, the alarm utility is shared across query
times, encouraging assistance requests on failure-labeled segments and
continued execution on success-labeled segments. The
intervention-feedback PPO term complements this outcome-based surrogate
with observed execution signals, including physical correction and
post-handback outcomes.


\section{Paired Recovery Cost of CALL and CALL+EGR}
\label{app:paired_recovery_cost}

We compare the recovery cost of CALL and CALL+EGR from matched
states. We use 40 states recorded at the first autonomous request,
covering 15 tasks across the four LIBERO suites. Each configuration
runs three repetitions per state, yielding 120 continuations per
method. The two configurations start from identical simulator,
controller, and $\pi_{0.5}$ random states. They share the frozen
offline failure-risk monitor at a common threshold and use Astra
with \texttt{medium} reasoning effort. CALL uses visual intervention
(Mode C in Fig.~\ref{fig:reasoning_action_paradigms}d). CALL+EGR also enables
reference-guided correction (Mode B), using bounded IK to approach
a visually compatible reference. Astra can use visual intervention
when no suitable reference is available. Both configurations allow
no-op handback to $\pi_{0.5}$ (Mode A).

We measure costs from the saved CALL state to task termination,
including subsequent requests. We count input and output tokens for
normal API requests and average costs over all continuations.
Effective time includes action execution, VLA inference, Astra
decision generation after connection, reference retrieval, and IK
computation. Effective throughput divides the total environment
steps after CALL, including steps executed by
$\pi_{0.5}$ after handback, by the corresponding total effective time.
The main-paper control frequency uses full-episode wall-clock time.
Paired bootstrap intervals resample tasks and matched states, with all
three repetitions of a state kept together.

\begin{table}[t]
\centering
\caption{Paired recovery cost from the saved CALL state to task
termination. Tokens, API requests, and effective time are reported per
continuation. Throughput uses the ratio of total steps to total
effective time.}
\label{tab:pica_matched_recovery_cost}
\resizebox{\columnwidth}{!}{%
\begin{tabular}{lrrrr}
\toprule
\textbf{Method}
& \textbf{Tokens (k) $\downarrow$}
& \textbf{API Requests $\downarrow$}
& \textbf{Effective Time (s) $\downarrow$}
& \textbf{Effective Steps/s $\uparrow$} \\
\midrule
CALL & 70.57 & 12.00 & 191.77 & 2.53 \\
CALL+EGR & 42.84 & 5.66 & 66.85 & 6.85 \\
\bottomrule
\end{tabular}%
}
\end{table}

CALL+EGR reduces mean token use by 39.3\% (paired 95\% CI:
17.4--56.3\%) and increases effective throughput by a factor of 2.71
(paired 95\% CI: [2.10, 3.62]) in Table~\ref{tab:pica_matched_recovery_cost}.
The first intervention consumes more tokens with CALL+EGR
(7.12k vs.\ 4.71k), but the mean number of API requests across the
continuation falls from 12.00 to 5.66. Direct visual correction steps
decrease from 36.01\% to 5.33\% of all executed steps. Bounded IK
corrections account for 37.84\% of the steps under CALL+EGR.

\end{document}